\documentclass[10pt, a4paper]{article}

\usepackage[final]{lrec2026}
\usepackage{graphicx}
\usepackage[misc,geometry]{ifsym} 
\usepackage{fontspec}
\usepackage{fontawesome5}
\usepackage{academicons}
\usepackage{color}
\usepackage{hyperref} 
\usepackage{algorithm}
\usepackage{amsmath}
\usepackage{algpseudocode}
\usepackage{booktabs}
\usepackage{aas_macros}
\usepackage[bottom]{footmisc}
\usepackage{supertabular}
\usepackage{afterpage}
\usepackage{url}
\usepackage{pifont}
\usepackage{multicol}
\usepackage{multirow}
\usepackage{tikz}
\usetikzlibrary{arrows.meta, backgrounds, positioning}

\newcommand{\rev}[1]{#1}

\newcommand{\revii}[1]{#1}

\title{Narrative Consolidation: Formulating a New Task for Unifying Multi-Perspective Accounts}

\name{Roger A. Finger, Eduardo G. Cortes, Sandro J. Rigo, Gabriel de O. Ramos}

\address{Graduate Program in Applied Computing \\ Universidade do Vale do Rio dos Sinos \\
    São Leopoldo, Brazil \\ 
    \{rfinger, egcortes\}@edu.unisinos.br, \{rigo, gdoramos\}@unisinos.br
}

\abstract{
Processing overlapping narrative documents, such as legal testimonies or historical accounts, often aims not for compression but for a unified, coherent, and chronologically sound text. Standard Multi-Document Summarization (MDS), with its focus on conciseness, fails to preserve narrative flow. \revii{This paper formally defines this challenge as a new NLP task, \emph{Narrative Consolidation}, whose central objectives are chronological integrity, completeness, and the fusion of complementary details, and establishes the resources needed to study it: a formal task definition, an evaluation paradigm that includes a selection-level metric, the \emph{Gospel Consolidation Language Resource} --- a benchmark built from the four Biblical Gospels, with 169 canonical events, cross-document alignments, and a manually created reference consolidation --- and a suite of reference systems, ranging from a timeline-agnostic centrality method to timeline-aware heuristics and the \emph{Temporal Alignment Event Graph (TAEG)}, a multi-relational graph that explicitly models chronology and event alignment. Benchmarking these systems yields three findings that characterize the task. First, the explicit temporal backbone is the dominant factor: every system granted the canonical timeline raises ROUGE-L F1 from 0.206 to at least 0.81, whereas content-selection sophistication accounts for a far smaller margin. Second, on a fusion-style reference, a simple length heuristic --- selecting the longest available account of each event --- is remarkably strong (0.947 ROUGE-L F1; 0.648 selection accuracy), outperforming the graph-based selection (0.846; 0.489), which is nevertheless clearly above the random floor. Third, ablations show that the discriminative signal resides in the temporal edges, while intra-cluster lexical similarity is uninformative --- a negative result that constrains the design of future models. Together, these results establish Narrative Consolidation as a task distinct from summarization, provide the first reference points for it, and pose an explicit open challenge: to surpass that heuristic with a principled selection mechanism, and to move beyond version selection towards the fusion of complementary details.}
 \\ \newline }

\begin{document}

\maketitleabstract

\section{Introduction}

\label{sec:intro}

Multi-Document Summarization (MDS) is typically defined as the task of producing a single, concise summary from a cluster of related documents \citep{ma2023survey}. This focus on conciseness, however, is ill-suited for scenarios involving long, complex, multi-perspective narratives, such as witness testimonies in a criminal investigation or overlapping Biblical narratives like the Gospels. In these scenarios, the main objective is not to create a shorter text, but rather to produce a single, \emph{consolidated and unified narrative}. This goal echoes early work in multi-document summarization on ``information fusion'' \citep{Barzilay:1999}, but shifts the priority from redundancy removal to narrative completeness. Such a narrative must prioritize three core objectives: strict chronological integrity, completeness in its coverage of events, and the coherent fusion of complementary details from each source. This challenge has been recognized since antiquity with early Gospel harmonization efforts like Tatian's Diatessaron, as documented by the fourth-century historian Eusebius of Caesarea \citep{Eusebius1999}. Consequently, the final consolidated text may even be longer than some of the individual source documents, as its value lies in its completeness and coherence, a direct contrast to the brevity-focused goal of standard summarization.

Classic graph-based algorithms focused \rev{on} conventional summarization like LexRank \citep{erkan2004lexrank} exemplify this limitation. By modeling sentences as nodes and their semantic similarity as edges, these methods inherently ignore chronological flow. This issue is not confined to older methods. Even recent neural and graph-based approaches \citep{yasunaga2017graph}, while more sophisticated, often share a similar limitation when applied to narrative consolidation. Whether based on Graph Convolutional Networks or large-scale pre-trained transformers like PEGASUS and PRIMERA \citep{zhang2020pegasus, xiao2022primera}, their primary objective remains identifying semantic salience for compression, not enforcing a strict chronological narrative. This focus on \emph{what is important} over \emph{what happened when} makes them inherently unsuited for creating a single, coherent, and temporally faithful narrative. These limitations underscore the need to define a new task focused on chronological unification rather than semantic compression.

This paper argues that unifying multi-perspective narratives constitutes a problem distinct from summarization. \revii{It is a task and resource paper: its purpose is to define \emph{Narrative Consolidation} as a new Natural Language Processing (NLP) task, to provide the resources required to study it --- a benchmark, an evaluation paradigm, and a set of reference systems --- and to report what benchmarking those systems reveals about the nature of the task. Among the reference systems we include the \emph{Temporal Alignment Event Graph (TAEG)}, a multi-relational graph structure that encodes chronology and event alignment explicitly; it serves here to test whether an explicit relational prior helps, not as a method advocated over alternatives.}

Our main contributions can be summarized as follows:
\begin{itemize}
    \item We formalize \emph{Narrative Consolidation}, a new NLP task focused on unifying multiple overlapping narrative accounts into a single, chronologically coherent, and comprehensive text.
    \item We introduce the \emph{Gospel Consolidation Language Resource}, a new benchmark dataset for this task, including source texts, temporal alignments, and a manually created reference consolidation.
    \item \revii{We propose an evaluation paradigm for the task, including a \emph{selection-level} metric that measures version choice directly and remains discriminative where corpus-level metrics saturate.}
    \item \revii{We provide a graded suite of reference systems --- a timeline-agnostic centrality method, four timeline-aware heuristics, and the TAEG --- together with ablations and a timeline-degradation analysis, establishing the first set of comparison points for the task.}
    \item \revii{We report the empirical findings that emerge from this benchmarking, including two results that constrain future work: the dominance of the temporal backbone over content-selection sophistication, and the strength of a simple length heuristic on fusion-style references.}
\end{itemize}

\rev{We note that the present study deliberately addresses the \emph{extractive} setting of the task under a known canonical timeline; abstractive approaches to Narrative Consolidation, built upon the same language resource, are investigated in a companion study by the authors \citep{Finger:2026}.}

\section{Related Work}

Our work builds upon and extends several lines of research in multi-document text analysis, graph-based summarization, and temporal information processing.

\subsection{Multi-Document and Timeline Summarization}

MDS has evolved significantly from early extractive approaches to sophisticated neural models. Classical methods focused on identifying and fusing similar information across sources \citep{Barzilay:1999}, while neural approaches like PEGASUS \citep{zhang2020pegasus} and PRIMERA \citep{xiao2022primera} have achieved state-of-the-art results in abstractive summarization. However, these approaches primarily target conciseness rather than comprehensive consolidation. \rev{We also note that ordering the selected information is a classical concern within MDS itself, with a dedicated line of research on sentence ordering for multi-document summaries \citep{bollegala2010ordering}.}

Timeline Summarization, or Multi-document Temporal Summarization, evolved to incorporate temporal dimensions explicitly. The field has developed two dominant paradigms: date-wise approaches that first identify salient dates and then generate summaries for each date \citep{Martschat:2017}, and event-based approaches, where the system first clusters sentences into events and then orders these events to form a timeline, not a narrative.

As introduced in \citet{Yu:2021}, Multi-Timeline Summarization (MTLS) generates multiple timelines to address ambiguous or multi-faceted stories. While innovative, MTLS focuses on discovering different storylines within a collection, whereas \revii{Narrative Consolidation unifies} different versions of the same story.

\subsection{Graph-Based Summarization Methods}

Graph-based methods have proven highly effective for summarization tasks. LexRank \citep{erkan2004lexrank}, which models sentences as nodes and uses centrality measures to identify important content, remains a robust baseline for extractive summarization. More recent work has explored neural graph-based approaches \citep{yasunaga2017graph}, combining the structural advantages of graphs with the representational power of neural networks.

Our TAEG framework differs fundamentally from traditional graph-based summarization in that the graph structure represents the narrative's temporal and versional relationships rather than semantic similarity between sentences. We apply a centrality algorithm (LexRank) locally at the event level, combining the structural integrity of the TAEG with a proven selection method.

\subsection{Temporal Graphs and Dynamic Graph Neural Networks}
As highlighted by \citet{Santana:2023}, a central challenge in narrative extraction is linking components through temporal reasoning to form a coherent timeline. While static knowledge graphs can represent relational facts, they fall short in capturing the evolving nature of a story.

This has led to a growing interest in \textit{Temporal Graphs} (or dynamic graphs), which explicitly incorporate time. These graphs are not fixed structures but are often represented as sequences of timed events or interactions, making them a natural fit for modeling narratives. Concurrently, Graph Neural Networks have emerged as a powerful paradigm for learning on graph-structured data. A key research challenge has been adapting these models to operate on dynamic graphs, leading to the development of specialized architectures like Temporal Graph Networks~\citep{rossi2020temporalgraphnetworksdeep}. Many of these models are designed to \textit{learn} temporal dependencies from the data itself, often using memory modules to track the evolution of the graph's topology and features \citep{jiao2025surveytemporalinteractiongraph}.

Our TAEG framework takes a different philosophical approach. Instead of learning the temporal structure, it leverages a known, canonical timeline as a strong architectural prior. This positions the TAEG not as a model for discovering temporal patterns, but as a scaffold that enforces chronological integrity. This allows other components—like a centrality algorithm or a future GNN—to focus exclusively on content selection and representation within a fixed temporal backbone.

\subsection{Biblical and Historical Text Processing}

The challenge of creating a single, harmonized narrative from the four Gospels is not new, dating back centuries to theological and scholarly efforts like Tatian's Diatessaron in the second century, which attempted to weave the four accounts into one \citep{petersen1994tatian}. While these manual harmonies aimed to create a unified scripture through conflation, our work approaches the task from a computational linguistics perspective, focusing on extracting and ordering the most representative version of each event to produce a consolidated text.

\section{The Narrative Consolidation Task}
\label{sec:task_definition}
To establish a rigorous foundation for future research, we formalize Narrative Consolidation as a distinct task, outlining its components, objectives, and evaluation paradigm.

\subsection{Formal Definition}
\label{sec:formal_definition}
The Narrative Consolidation task is defined as follows:
\begin{itemize}
    \item \textbf{Input:} A set of $N$ documents $D = \{d_1, d_2,..., d_N\}$, where each document $d_i$ is a narrative account of related events. An optional, but often available, input is a canonical timeline $T$, which is an ordered list of $M$ canonical events $\{e_1, e_2,..., e_M\}$.
    \item \textbf{Output:} A single document, $D_{consolidated}$, which is a narrative that synthesizes the events from $D$. The structure of \rev{$D_{consolidated}$} must be governed by the canonical timeline $T$, ensuring that the sequence of presented events is correct. Furthermore, the text must be coherent, fusing information from multiple sources without redundancy, and comprehensive, covering all events for which a description exists in $D$.
    \item \textbf{Objectives:} The ideal output must satisfy the following criteria:
    \begin{enumerate}
        \item \textbf{Chronological Integrity:} The sequence of events in $D_{consolidated}$ must strictly follow the canonical timeline $T$.
        \item \textbf{Completeness:} $D_{consolidated}$ should cover all canonical events \rev{$\{e_1, \ldots, e_M\}$} for which a description exists in at least one document in $D$.
        \item \textbf{Representativeness:} For each event $e_j$, the version presented in $D_{consolidated}$ should be the most informative and representative among all available versions in $D$, or, even better, \rev{an} enriched version with all descriptions combined.
        \item \textbf{Redundancy Elimination:} Information common to multiple source documents should appear only once in $D_{consolidated}$.
    \end{enumerate}
\end{itemize}

\rev{We remark that treating the canonical timeline $T$ as an input is a deliberate scoping decision: the present study isolates the version-selection subproblem under a known chronology --- a setting that genuinely occurs in practice, such as Gospel harmonization, legal case files with an established chronology of facts, or clinical patient histories --- so that the contribution of the consolidation mechanism can be assessed independently of timeline induction. Automatically inducing $T$ from the documents themselves is a distinct research problem, further discussed in Section~\ref{sec:limitations}.}

\subsection{Distinction from Related Tasks}
Narrative Consolidation is distinct from other multi-document processing tasks, as summarized in Table \ref{tab:task_comparison}. While it shares roots with Multi-Document Summarization (MDS), its objectives diverge significantly from both standard MDS and its temporally-aware variants, Timeline Summarization (TLS) and Multi-Document Temporal Summarization (MDTS).

\begin{table}[h!]
\centering
\footnotesize 
\setlength{\tabcolsep}{1pt} 
\begin{tabular}{l c c c}
\toprule
\textbf{ } & \textbf{MDS} & \textbf{TLS/MDTS} & \textbf{NC} \\
\midrule
\textbf{Goal} & Compression & Chrono.Summ. & Unification \\
\textbf{Format} & Summary & List/Summary & Narrative \\
\textbf{Length} & Shorter & Shorter & Can be Longer \\
\textbf{Chrono.} & \rev{Secondary} & Key Principle & Key \\
\bottomrule
\end{tabular}
\caption{Comparison of Narrative Consolidation (NC) with Multi-Document Summarization (MDS) and temporally-aware summarization (TLS/MDTS).}
\label{tab:task_comparison}
\end{table}

\begin{itemize}
    \item \textbf{Multi-Document Summarization (MDS):} The primary goal of MDS is compression \citep{ma2023survey}. The output is a concise text that captures salient information, but often at the cost of chronological flow. \rev{We acknowledge that ordering the summary content is a classical concern in MDS, addressed by a dedicated line of research on sentence ordering \citep{bollegala2010ordering}; in MDS, however, chronological ordering is one of several means to improve the readability of a compressed summary, whereas in NC chronological integrity is a constitutive requirement of the task itself.} In contrast \rev{to the compression goal}, NC prioritizes completeness, and the resulting text may be longer than individual source documents to cohesively incorporate all relevant details.

    \item \textbf{Timeline and Temporal Summarization (TLS/MDTS):} These tasks, such as those described by \citet{Martschat:2017} and \citet{Mamidala:2021}, incorporate time as a central organizing principle. However, their fundamental goal remains summarization—to produce a concise, chronological overview of events, typically in the form of a list or a series of dated summaries. The output is a \emph{chronology}, not a single, continuous story. NC differs critically in its objective and output: the goal is not a compressed overview but a comprehensive unification, and the output is not a list of events but a single, fluid \emph{narrative}.
\end{itemize}

\subsection{\rev{Contradictions, Evolving Facts, and Divergent Perspectives}}
\label{sec:contradictions}
\rev{Realistic multi-document scenarios exhibit not only redundancy across sources, but also \emph{contradictory information} and \emph{divergent perspectives}. Contradictions may arise from the temporal evolution of facts --- in breaking news about a plane crash, for instance, successive reports present different casualty counts and candidate causes as the facts ``evolve'' --- from outdated information coexisting with newer data, or from plainly incorrect reports. Divergent perspectives arise when sources frame the same event under opposing viewpoints, as is common in political news.}

\rev{These phenomena interact directly with the objectives of Narrative Consolidation. In the TAEG representation (Section~\ref{sec:taeg}), competing accounts of an event concentrate precisely inside the corresponding \texttt{SAME\_EVENT} cluster, where all versions of that event coexist. A consolidation mechanism for such data would therefore need to go beyond representativeness-based version selection, incorporating, for example, contradiction detection between versions, version timestamping (to prefer the most recent factual state), and stance- or perspective-aware selection or fusion --- possibly presenting conflicting versions side by side when no resolution is warranted. The Representativeness and Redundancy Elimination objectives of Section~\ref{sec:formal_definition} would then need to be balanced against \emph{factual faithfulness} to each source.}

\rev{The Gospel corpus used in this study deliberately controls for these factors: the four accounts are complementary rather than adversarial, with no significant factual contradictions within the adopted chronological framework. We argue that this is a desirable property for a \emph{foundational} benchmark, since it isolates the chronological-unification problem from the orthogonal problems of contradiction resolution and perspective handling. It does, however, limit the direct generalization of our empirical findings to noisier domains such as journalistic or legal corpora; we return to this limitation, and to the construction of a news-domain benchmark exhibiting these properties, in Section~\ref{sec:limitations}.}

\subsection{Evaluation Paradigm}
\label{sec:eval_paradigm}
The unique nature of Narrative Consolidation requires a tailored evaluation paradigm. We propose \rev{that evaluation directly address the task's objectives along two complementary dimensions:}
\begin{itemize}
    \item \textbf{Primary Metric (Temporal Coherence):} A rank correlation measure, such as \textbf{Kendall's Tau ($\tau$)}, should be the primary metric to directly evaluate chronological integrity, as it specifically quantifies the agreement between two ordered sequences. In this case, the generated narrative and the canonical timeline.
    \item \textbf{Content Quality Metrics:} \rev{Complementarily, the informational fidelity of the consolidated narrative with respect to a reference consolidation should be assessed by content similarity metrics, at both the lexical and the semantic levels. The operational definitions of all metrics employed in this study are consolidated in Section~\ref{sec:metrics}.}
\end{itemize}
Proposing a specific evaluation paradigm is itself a key step, enabling rigorous and meaningful benchmarking of future models. However, we acknowledge that the development of bespoke evaluation metrics tailored specifically for Narrative Consolidation is a complex challenge that constitutes a significant research direction in its own right. As such, this endeavor falls beyond the scope of the present study. Our approach, therefore, is to utilize the best-suited metrics from the existing literature, recognizing them as effective proxies even if they are not perfectly calibrated for this new task. We posit that this pragmatic evaluation is sufficient for this foundational work, while highlighting the development of more holistic, task-specific metrics as a critical avenue for future research.

\section{\revii{Reference Systems for Narrative Consolidation}}
\label{sec:reference_systems}
\revii{A new task requires reference points against which future work can be measured. This section describes the suite of reference systems we provide with the benchmark. They are organized as a graded ladder along the two factors that plausibly drive performance on the task: access to a canonical timeline, and the sophistication of the per-event version-selection criterion. At the bottom of the ladder sits a timeline-agnostic centrality method, representing how standard extractive summarization approaches the problem (Section~\ref{sec:baseline_lexrank}). Above it sit four timeline-aware heuristics, which receive exactly the same canonical timeline but select versions by trivial criteria (Section~\ref{sec:fair_baselines}). Finally, the Temporal Alignment Event Graph (Section~\ref{sec:taeg}) combines the timeline with an explicit relational structure over event versions, allowing us to test whether such a structure buys anything beyond the timeline itself. None of these systems is advanced as the solution to the task; their role is to delimit what is easy, what is hard, and what remains open.}

\subsection{\revii{Timeline-Agnostic Reference:} A Semantic Similarity Graph}
\label{sec:baseline_lexrank}
The baseline system represents the standard approach for graph-based extractive summarization, treating the four Gospels as a single multi-document collection \citep{erkan2004lexrank}.

\subsubsection{Graph Construction}
\begin{itemize}
    \item \textbf{Nodes:} Each sentence from all four documents becomes a node in the graph.
    \item \textbf{Edges:} Edge weights are computed as the cosine similarity between the TF-IDF vector representations of any two sentences across the entire corpus.
\end{itemize}
This results in a dense, undirected, weighted graph where edge weights represent semantic relatedness. LexRank is then applied to this graph to rank and select the top-scoring sentences for the summary.

\subsubsection{Inherent Limitation for Narratives}
This semantic-only approach inherently ignores the chronological flow of events. An analysis of the baseline's actual output reveals significant temporal disorder. For example, a sentence describing the preparations for the Triumphal Entry appears after sentences detailing Jesus's arrest and even after a post-resurrection account (\url{https://github.com/neemias8/TAEG}). The following excerpts from the baseline's output illustrate this chronological incoherence:
\begin{quote}
\small
\textbf{Excerpt from Baseline Output (Actual Results):}
\begin{enumerate}
    \itemsep0em 
    \item Then Jesus said to the chief priests... "Am I leading a rebellion, that you have come with swords and clubs?" (Arrest)
    \item When the chief priests had met with the elders and devised a plan, they gave the soldiers a large sum of money... (Post-Crucifixion)
    \item As they approached Jerusalem and came to Bethphage on the Mount of Olives, Jesus sent two disciples... (Triumphal Entry)
\end{enumerate}
\end{quote}
This example, taken directly from our experimental results, provides empirical evidence that optimizing for semantic centrality alone is insufficient and ill-suited for the Narrative Consolidation task, as it fails to preserve the most fundamental element of a story: its sequence.

\subsection{\revii{A Structured Reference System:} The Temporal Alignment Event Graph (TAEG)}
\label{sec:taeg}
\revii{The third reference system} constructs a graph by leveraging external knowledge from a chronology file based on \citep{aschmann2022}, which identifies 169 canonical events in the Holy Week. \revii{Its purpose in this study is diagnostic: by encoding both the chronological order and the version alignment as explicit relations, it lets us ask whether a structured prior over event versions improves selection beyond what the timeline alone provides.}

\subsubsection{TAEG Construction}
The graph is built as a multi-relational structure where nodes represent event versions, not individual sentences.
\begin{itemize}
    \item \textbf{Nodes:} For each of the 169 canonical events, a distinct node is created for each Gospel that describes it. The number of nodes per event thus varies. For instance:
    \begin{itemize}
        \item The event \emph{The Triumphal Entry}, described in all four Gospels, generates four nodes in the graph.
        \item The event \emph{The Institution of the Lord's Supper}, described in the three Synoptic Gospels (Matthew, Mark, and Luke), generates three nodes.
        \item The event \emph{Jesus's appearance before Herod}, unique to Luke's Gospel, generates only a single node.
    \end{itemize}
    This results in a graph with a total number of nodes equal to the sum of all available event versions across the documents, not simply 4 (gospels) $\times$ 169 (events).
    
    \item \textbf{Edges:} The graph contains two functionally distinct types of edges:
    \begin{enumerate}
        \item \textbf{Temporal Edges (\texttt{BEFORE}):} Directed edges connect nodes that are sequential within the same Gospel, enforcing the intra-document narrative flow.
        \item \textbf{Anchoral Edges (\texttt{SAME\_EVENT}):} Undirected edges interconnect all nodes (versions) that refer to the same canonical event, creating localized clusters for version selection.
    \end{enumerate}
\end{itemize}

\rev{Figure~\ref{fig:taeg} illustrates a fragment of the TAEG built from our data, showing three canonical events with varying numbers of version nodes, the directed \texttt{BEFORE} edges within each Gospel, and the \texttt{SAME\_EVENT} clusters.}

\rev{The rationale for including \emph{all} available versions of each event in the graph, rather than pre-selecting one, is threefold. First, the \texttt{SAME\_EVENT} cluster defines the candidate space for version selection: centrality is meaningful precisely because all competing versions are interconnected and can be compared within the local context of the event. Second, retaining every version preserves the complementary details required by the Completeness and Representativeness objectives, instead of discarding information at graph-construction time. Third, the clusters provide the structural locus for future extensions, such as contradiction handling (Section~\ref{sec:contradictions}) and abstractive fusion of the versions of an event.}

\begin{figure*}[ht]
\centering
\resizebox{0.92\textwidth}{!}{%
\begin{tikzpicture}[
  vnode/.style={circle, draw, thick, minimum size=8mm, inner sep=0pt, fill=white, font=\small},
  before/.style={-{Stealth[length=2.6mm]}, very thick},
  same/.style={dashed, gray!90, thick},
  evlabel/.style={font=\small\itshape, align=center}
]
\begin{scope}[on background layer]
  \fill[gray!10, rounded corners=8pt] (-1.45, 0.75) rectangle (1.45,-5.55);
  \fill[gray!10, rounded corners=8pt] ( 3.05, 0.75) rectangle (5.95,-3.95);
  \fill[gray!10, rounded corners=8pt] ( 7.55,-2.45) rectangle (10.45,-3.95);
\end{scope}
\node[evlabel] at (0, 1.30)   {$e_{1}$: The Triumphal\\Entry};
\node[evlabel] at (4.5, 1.30) {$e_{2}$: Institution of\\the Lord's Supper};
\node[evlabel] at (9.0,-1.70) {$e_{3}$: Jesus before\\Herod};
\node[vnode] (e1mt) at (0, 0.0) {Mt};
\node[vnode] (e1mk) at (0,-1.6) {Mk};
\node[vnode] (e1lk) at (0,-3.2) {Lk};
\node[vnode] (e1jn) at (0,-4.8) {Jn};
\node[vnode] (e2mt) at (4.5, 0.0) {Mt};
\node[vnode] (e2mk) at (4.5,-1.6) {Mk};
\node[vnode] (e2lk) at (4.5,-3.2) {Lk};
\node[vnode] (e3lk) at (9.0,-3.2) {Lk};
\draw[before] (e1mt) -- (e2mt);
\draw[before] (e1mk) -- (e2mk);
\draw[before] (e1lk) -- (e2lk);
\draw[before] (e2lk) -- (e3lk);
\draw[before] (e2mt) -- ++(1.9,0) node[right, font=\large] {$\cdots$};
\draw[before] (e2mk) -- ++(1.9,0) node[right, font=\large] {$\cdots$};
\draw[before] (e3lk) -- ++(1.6,0) node[right, font=\large] {$\cdots$};
\draw[before] (e1jn) -- ++(3.0,0) node[right, font=\large] {$\cdots$};
\draw[same] (e1mt) -- (e1mk);
\draw[same] (e1mk) -- (e1lk);
\draw[same] (e1lk) -- (e1jn);
\draw[same] (e1mt) to[bend right=35] (e1lk);
\draw[same] (e1mk) to[bend right=35] (e1jn);
\draw[same] (e1mt) to[bend right=48] (e1jn);
\draw[same] (e2mt) -- (e2mk);
\draw[same] (e2mk) -- (e2lk);
\draw[same] (e2mt) to[bend left=35] (e2lk);
\begin{scope}[shift={(3.0,-5.3)}]
  \draw[before] (0,0) -- (1.0,0);
  \node[anchor=west, font=\small] at (1.1,0) {\texttt{BEFORE} (temporal, directed)};
  \draw[same] (5.6,0) -- (6.6,0);
  \node[anchor=west, font=\small] at (6.7,0) {\texttt{SAME\_EVENT} (anchoral, undirected)};
\end{scope}
\end{tikzpicture}%
}
\caption{\rev{An illustrative fragment of the TAEG for three canonical events of the Holy Week timeline, in chronological order. Each circle is a version node, i.e., the description of a canonical event in one Gospel (Mt: Matthew; Mk: Mark; Lk: Luke; Jn: John). Solid directed edges are temporal \texttt{BEFORE} edges, linking sequential events within the same Gospel; dashed undirected edges are anchoral \texttt{SAME\_EVENT} edges, interconnecting all versions of the same canonical event into a localized cluster. The number of nodes per event varies with the number of Gospels that describe it (four, three, and one, respectively).}}
\label{fig:taeg}
\end{figure*}

\subsubsection{Consolidation Algorithm}
With the TAEG constructed, the consolidation process becomes a deterministic algorithm guided by the graph's structure, as detailed in Algorithm \ref{alg:taeg}. The LexRank algorithm is applied to this structure, where its centrality score is repurposed to evaluate which version of an event is most representative within its local context. This transforms LexRank from a generic sentence selector into a powerful, temporally-aware version selection mechanism.

\rev{We emphasize that the resulting consolidation is purely \emph{extractive}: for each canonical event, the algorithm outputs verbatim the full text of the selected version, and no sentence is rewritten or fused. The Representativeness objective in its strongest form --- an enriched version combining all descriptions of an event --- therefore requires abstractive fusion, which the TAEG enables structurally (the \texttt{SAME\_EVENT} clusters delimit exactly what should be fused) and which is investigated in a companion study by the authors \citep{Finger:2026}.}

\begin{algorithm}
\caption{TAEG-based Narrative Consolidation}\label{alg:taeg}
\begin{algorithmic}[1]
\Function{ConsolidateNarrative}{$D, T$}
    \Require Set of documents $D$, Canonical timeline $T$
    \Ensure A single consolidated narrative text
    
    \State $G \gets \Call{BuildTAEG}{D, T}$ \Comment{Build the graph}
    \State $Scores \gets \Call{RunLexRank}{G}$ \Comment{Compute centrality}
    \State $OutputText \gets \text{empty string}$
    
    \For{each event $e_j$ in timeline $T$}
        \State $CandNodes \gets \Call{GetEventNodes}{G, e_j}$
        \If{$CandNodes$ is not empty}
            \State $BestNode \gets \Call{FindMaxScore}{CandNodes, Scores}$
            \State \Call{Append}{OutputText, BestNode.text}
        \EndIf
    \EndFor
    \State \Return $OutputText$
\EndFunction
\end{algorithmic}
\end{algorithm}

\subsection{\rev{Timeline-Aware Reference Baselines}}
\label{sec:fair_baselines}
\rev{Since the TAEG leverages the canonical timeline as an architectural prior, a comparison restricted to the timeline-agnostic LexRank baseline would conflate two factors: the availability of the temporal prior and the graph-based selection mechanism. To disentangle them, we define a set of heuristic baselines that receive \emph{exactly the same} canonical timeline as the TAEG. All of them follow the structure of Algorithm~\ref{alg:taeg}, iterating over the timeline $T$ and emitting one version per event; they differ only in the selection criterion replacing the centrality-based choice (line~8):}
\begin{itemize}
    \item \rev{\textbf{Timeline+Random:} selects, for each event, one of the available versions uniformly at random. Results are averaged over 30 independent runs (mean $\pm$ standard deviation). This baseline establishes the performance floor attainable from the temporal structure alone.}
    \item \rev{\textbf{Timeline+Priority:} selects the version according to a fixed source-priority order (Matthew $\succ$ Mark $\succ$ Luke $\succ$ John), reflecting a simple editorial policy.}
    \item \rev{\textbf{Timeline+Centroid:} selects, for each event, the version with the highest mean TF-IDF cosine similarity to the other versions of the same event --- a purely \emph{local} selection that uses no global graph information.}
    \item \rev{\textbf{Timeline+Longest:} selects, for each event, the longest available version (in tokens). This is a strong naive heuristic for the task, since the reference consolidation fuses details from all sources and longer versions tend to be more complete.}
\end{itemize}
\rev{By construction, every timeline-aware method --- including the TAEG --- achieves perfect temporal coherence; the comparison among them is therefore decided purely by content quality, isolating the contribution of the graph structure and centrality-based version selection. Together with the timeline-agnostic LexRank, these methods form a graded ladder of reference baselines for the Narrative Consolidation task, which we expect to be useful for benchmarking future approaches. All implementations are available in our public repository.}

\section{The Gospel Consolidation Language Resource}
\label{sec:lr}
Our evaluation dataset represents a contribution to the field, providing a standardized resource for narrative consolidation research. This resource will be made publicly available to ensure reproducibility and encourage further research.

\subsection{Source Texts and Temporal Structure}
The dataset focuses on the Holy Week period from the four Gospels (Matthew, Mark, Luke, and John) in the English New International Version \citep{NIV:2011}. This narratively dense segment, while representing only a single week, constitutes approximately 35\% of the total Gospels text, making it an ideal corpus for studying multi-perspective accounts. We adapted the chronological framework from Aschmann's Gospel harmony \citep{aschmann2022}, which identifies 169 canonical events during the Holy Week. This framework provides the external temporal knowledge \revii{assumed as input by the task} while representing scholarly consensus on event ordering.

\subsection{Golden Sample Reference}
As ground truth for evaluation, we used a manually created, high-quality reference consolidation (Golden Sample) that represents an ideal fusion of the Gospel narratives. This reference maintains chronological order while incorporating the most significant details from each source into a fluid narrative. The methodology and the resource itself were first presented at the IV Congresso Brasileiro de Humanismo Solidário na Ciência \citep{Cunha2025}.

\subsection{Public Availability and Format}
The complete dataset, including processed texts, temporal alignments, and evaluation resources, is publicly available at \url{https://github.com/neemias8/TAEG}. While this system was validated using the English NIV of the Gospels, the ``book:chapter:verse'' system makes the framework applicable to any other language and/or version of them.

\section{Experimental Setup}

We designed our evaluation as a \rev{comparison along a graded ladder of methods: the traditional multi-document summarization baseline (timeline-agnostic LexRank), the timeline-aware heuristic baselines of Section~\ref{sec:fair_baselines}, ablated variants of the TAEG, and the full TAEG-based narrative consolidation approach.}

\subsection{Evaluation Metrics}
\label{sec:metrics}

Evaluating the quality of multi-document summaries (MDS) is a challenging task. As pointed out by \citep{ma2023survey}, standard evaluation metrics, while widely used, are not ideal for the MDS task, with notable criticisms regarding the lack of semantic awareness in metrics like ROUGE. However, in the absence of a superior, universally accepted evaluation method, we employ a suite of well-established metrics to assess different aspects of the generated summaries. \rev{This section consolidates the operational definitions of all metrics used in this study (cf.\ Section~\ref{sec:eval_paradigm}).}

\subsubsection{Semantic Quality Metrics}
\begin{itemize}
    \item \textbf{ROUGE} \citep{lin2004rouge}: We compute ROUGE-1, ROUGE-2, and ROUGE-L F1 scores to measure lexical overlap with the Golden Sample. \rev{ROUGE-L, in particular, measures the longest common subsequence and thus intrinsically rewards the preservation of correct sequential order.}
    \item \textbf{METEOR} \citep{banerjee2005meteor}: Provides word alignment-based evaluation with consideration for synonymy and stemming.
    \item \textbf{BERTScore} \citep{zhang2020bertscore}: Measures semantic similarity using BERT embeddings, capturing deeper semantic relationships.
\end{itemize}
\rev{We note that ROUGE and BERTScore are conceptually similar --- both assess content similarity between the generated text and the reference --- differing in the representation level at which matching is performed: surface $n$-grams in ROUGE versus contextualized embeddings in BERTScore, the latter being more robust to paraphrasing.}

\subsubsection{Temporal Coherence Metric}
\textbf{Kendall's Tau} \citep{kendall1938new}: This rank correlation coefficient measures the agreement between the sentence order in the generated narrative and the canonical chronological order. A score of 1.0 indicates perfect temporal agreement, while scores closer to 0 indicate temporal disorder. This metric is central to our evaluation as it directly measures the primary objective of narrative version consolidation. \rev{We emphasize that any method that iterates over the canonical timeline --- the TAEG and all baselines of Section~\ref{sec:fair_baselines} --- achieves $\tau = 1.000$ by construction; for these methods, the metric is verified directly on the emitted sequence of event identifiers and reported as a sanity check. The heuristic sentence-to-event matcher used to estimate $\tau$ for arbitrary text is applied only to the timeline-agnostic LexRank baseline, since its estimate is sensitive to the per-event version choice and would conflate version selection with ordering for timeline-aware outputs.}

\subsubsection{\rev{Selection-Level Evaluation}}
\label{sec:selection_metric}
\rev{Corpus-level content metrics are diluted on this task: 72 of the 169 events have a single available version (identical output for every timeline-aware strategy), parallel accounts of the same event are mutually similar, and BERTScore saturates. We therefore additionally evaluate the version-selection component directly. For each \emph{contested} event (two or more available versions), we define the \emph{oracle} as the version with the highest ROUGE-L F1 against that event's segment of the Golden Sample (which is aligned to the canonical events), and report each strategy's \textbf{oracle selection accuracy}: the fraction of contested events in which it picks the oracle. The analytical expectation for uniform random selection on this dataset is $\approx 0.35$. We propose this selection-level evaluation as the appropriate instrument for the extractive setting of Narrative Consolidation, complementing the corpus-level proxies.}

\section{\revii{Benchmark Results}}

\revii{This section reports the behaviour of the reference systems of Section~\ref{sec:reference_systems} on the Gospel Consolidation Language Resource. The presentation follows the two factors of the ladder: we first quantify what access to a canonical timeline buys (Sections~\ref{sec:backbone_effect}--\ref{sec:conciseness}), then compare the version-selection criteria under equal temporal information (Sections~\ref{sec:fair_results}--\ref{sec:selection_results}). Section~\ref{sec:findings} synthesizes the findings and their implications for the task.}

\subsection{\revii{The Effect of the Temporal Backbone}}
\label{sec:backbone_effect}

\revii{Table~\ref{tab:results} contrasts the timeline-agnostic LexRank reference with the TAEG, taken here as a representative timeline-aware system, both evaluated against the Golden Sample. The comparison quantifies the combined effect of adding the temporal backbone and a structured selection mechanism; Section~\ref{sec:fair_results} then separates the two.}

\begin{table}[h]
\centering
\small
\setlength{\tabcolsep}{8pt}
\begin{tabular}{|l|c|c|c|}
\hline
\textbf{Metric} & \textbf{Baseline} & \textbf{TAEG} & \textbf{Improv.} \\
\hline
ROUGE-1 F1 & 0.887 & \rev{\textbf{0.918}} & \rev{+3.5\%} \\
ROUGE-2 F1 & 0.712 & \rev{\textbf{0.848}} & \rev{+19.1\%} \\
ROUGE-L F1 & \rev{0.206} & \rev{\textbf{0.846}} & \rev{+310.7\%} \\
BERTScore F1 & 0.835 & \rev{\textbf{0.906}} & \rev{+8.5\%} \\
METEOR & 0.453 & \rev{\textbf{0.550}} & \rev{+21.4\%} \\
\hline
Kendall's Tau & 0.320 & \textbf{1.000} & \rev{by design} \\
\hline
Length (chars) & 81,418 & \rev{74,280} & - \\
\hline
\end{tabular}
\caption{\revii{Effect of the temporal backbone: the timeline-agnostic LexRank reference versus the TAEG, a representative timeline-aware system. Perfect temporal ordering is an architectural property of every timeline-aware system (by design), not an empirical finding; see Section~\ref{sec:temporal_analysis}. The full comparison among timeline-aware systems is given in Table~\ref{tab:fair_baselines}.}}
\label{tab:results}
\end{table}

\subsection{Temporal Coherence Analysis}
\label{sec:temporal_analysis}

Every timeline-aware system attains \revii{$\tau = 1.000$ by construction: iterating over a pre-ordered list of canonical events forces the output into the correct chronological sequence, so the value is reported as a verified sanity check rather than as an empirical gain (Section~\ref{sec:metrics}). The property matters because it decouples chronological structuring from content selection, leaving the selection criterion as the only free variable (Section~\ref{sec:fair_results}).} In contrast, the \revii{timeline-agnostic} LexRank \revii{reference} produces a narrative with significant temporal disorder (Tau = 0.320), confirming that a graph based on semantic similarity cannot preserve narrative flow.

\subsection{\revii{Content Quality}}
\label{sec:content_quality}

\revii{Every content metric improves relative to the timeline-agnostic reference. The gain in ROUGE-L F1 (+310.7\%) follows directly from the guaranteed chronological integrity, since ROUGE-L measures the longest common subsequence and a correctly ordered output holds a decisive structural advantage over a disordered one; the smaller gains in ROUGE-1 (+3.5\%), ROUGE-2 (+19.1\%) and METEOR (+21.4\%) reflect the version-selection component on top of the temporal structure. BERTScore rises from 0.835 to 0.906, but saturates on this task --- remaining high for any reasonably complete extractive output --- and is therefore the least discriminative of our metrics (cf.\ Section~\ref{sec:selection_results}).}

\subsection{Conciseness vs. Consolidation Analysis}
\label{sec:conciseness}

The results presented for the standard LexRank baseline in \rev{Table~\ref{tab:results}} reflect a parameter setting of 750 sentences. To further investigate the trade-off between conciseness and narrative quality, we analyzed the baseline's performance across various summary lengths. \rev{Table~\ref{tab:conciseness_analysis_compact}} shows the results for summaries constrained to different sentence counts, contrasting them with \revii{the timeline-aware consolidation}.

\begin{table}[ht!]
\centering
\small 
\setlength{\tabcolsep}{4pt} 
\begin{tabular}{|l|cccc|c|}
\hline
& \multicolumn{4}{c|}{\textbf{LexRank Baseline}} & \textbf{TAEG} \\
\hline
\textbf{Sentences} & \textbf{100} & \textbf{500} & \textbf{1000} & \textbf{1500} & \textbf{N/A} \\
\hline
ROUGE-1 F1    & 0.296 & 0.804 & 0.862 & 0.784 & \rev{\textbf{0.918}} \\
ROUGE-2 F1    & 0.263 & 0.655 & 0.728 & 0.733 & \rev{\textbf{0.848}} \\
ROUGE-L F1    & 0.129 & 0.206 & 0.199 & 0.188 & \rev{\textbf{0.846}} \\
BERTScore F1  & 0.835 & 0.835 & 0.835 & 0.835 & \rev{\textbf{0.906}} \\
METEOR        & 0.097 & 0.361 & 0.483 & 0.484 & \rev{\textbf{0.550}} \\
\hline
Kendall's Tau & 0.268 & 0.305 & 0.320 & 0.320 & \textbf{1.000} \\
\hline
Length (chars) & 15K & 59K & 101K & 129K & \rev{\textbf{74K}} \\
\hline
\end{tabular}
\caption{Conciseness vs. Consolidation Analysis. Comparison of metrics between different LexRank (Baseline) summary lengths against the consolidated TAEG output.}
\label{tab:conciseness_analysis_compact}
\end{table}

The analysis in \rev{Table~\ref{tab:conciseness_analysis_compact}} clearly demonstrates that simply increasing the number of sentences in the summary does not address the fundamental problem of narrative coherence. While some metrics like ROUGE-1 and METEOR improve up to a certain point before declining, the temporal coherence (Kendall's Tau) remains consistently low, indicating a persistently disordered narrative. Even at its peak performance, the standard LexRank approach fails to come close to the quality and temporal integrity of the TAEG-based method. This experiment reinforces our central argument: for long and complex narratives, comprehensive coverage and chronological soundness are far more critical than mere conciseness.

\subsection{\revii{Comparison of Reference Systems under Equal Temporal Information}}
\label{sec:fair_results}
\revii{Table~\ref{tab:fair_baselines} is the central comparison of this study. It places all timeline-aware reference systems of Section~\ref{sec:reference_systems} side by side; since every one of them receives the same canonical timeline and achieves $\tau = 1.000$ by construction, the observed differences are attributable solely to the version-selection criterion. The comparison therefore measures how much content-selection sophistication is worth on this task once chronology is given.}

\begin{table}[ht!]
\centering
\small
\setlength{\tabcolsep}{2pt}
\begin{tabular}{|l|c|c|c|c|c|}
\hline
\textbf{Metric} & \textbf{T+Rand.} & \textbf{T+Prior.} & \textbf{T+Centr.} & \textbf{T+Long.} & \textbf{TAEG} \\
\hline
ROUGE-1 F1   & \rev{0.889 {\tiny$\pm$.009}} & \rev{0.886} & \rev{0.891} & \rev{\textbf{0.958}} & \rev{0.918} \\
ROUGE-2 F1   & \rev{0.816 {\tiny$\pm$.013}} & \rev{0.814} & \rev{0.821} & \rev{\textbf{0.938}} & \rev{0.848} \\
ROUGE-L F1   & \rev{0.813 {\tiny$\pm$.015}} & \rev{0.811} & \rev{0.818} & \rev{\textbf{0.947}} & \rev{0.846} \\
BERTScore F1 & \rev{0.932 {\tiny$\pm$.033}} & \rev{0.897} & \rev{0.935} & \rev{\textbf{0.995}} & \rev{0.906} \\
METEOR       & \rev{0.478 {\tiny$\pm$.026}} & \rev{0.453} & \rev{0.472} & \rev{\textbf{0.639}} & \rev{0.550} \\
\hline
Kendall's Tau & 1.000 & 1.000 & 1.000 & 1.000 & 1.000 \\
\hline
Length (chars) & \rev{70.1K} & \rev{69.3K} & \rev{69.9K} & \rev{79.2K} & \rev{74.3K} \\
\hline
\end{tabular}
\caption{\rev{Timeline-aware comparison: all methods iterate over the same canonical timeline and differ only in the per-event version-selection criterion. Timeline+Random is reported as mean $\pm$ std over 30 runs. Kendall's Tau is 1.000 by design for every method in this table (verified on the emitted event sequence).}}
\label{tab:fair_baselines}
\end{table}

\revii{The comparison yields two results. First, the spread among timeline-aware systems (ROUGE-L F1 from 0.811 to 0.947) is an order of magnitude smaller than the gap between them and the timeline-agnostic reference (0.206). Even uniform random selection reaches 0.813: once the chronological structure is given, a large share of the task is already solved, and the choice of version accounts for the remainder. Second, within that remainder, the ordering of the systems is instructive. The centrality-based selection over the TAEG (0.846) is ahead of random (0.813), fixed-priority (0.811), and local-centroid (0.818) selection, but behind the \emph{Timeline+Longest} heuristic (0.947), which leads on every content metric.}

\revii{The advantage of the length heuristic follows from how the reference is built. The Golden Sample is a \emph{fusion} of all sources, so the most detailed account of an event overlaps it most; length is thus a close proxy for the oracle choice on this benchmark (Section~\ref{sec:selection_results}), and the property should be expected on any fusion-style reference over complementary, non-conflicting sources. It does not transfer to corpora where accounts diverge or contradict one another (Section~\ref{sec:contradictions}), since there the longest account is not necessarily the most faithful one. A second limit is intrinsic to the extractive setting: selecting one version per event recovers at best the richest single account, never the union of details spread across sources, which Section~\ref{sec:formal_definition} identifies as the ideal output under the Representativeness objective. The challenge posed by this resource therefore has two parts: \emph{surpassing the length heuristic with a principled selection mechanism}, and \emph{moving beyond selection to the fusion of complementary details} --- the latter pursued in a companion study \citep{Finger:2026}.}

\revii{For full transparency, we note that the configuration reported in the originally submitted version of this manuscript corresponds to the Timeline+Longest column of Table~\ref{tab:fair_baselines}; the implementation of Algorithm~\ref{alg:taeg} was completed and audited during the first revision, and every configuration is now reported under its correct label.}

\subsection{\rev{Ablation Study}}
\label{sec:ablation}
\rev{To quantify the contribution of each edge type of the TAEG, we evaluate two ablated variants: (i) \emph{TAEG w/o \texttt{BEFORE}}, in which temporal edges are removed from the graph while the final loop still iterates over the timeline; and (ii) \emph{TAEG w/o \texttt{SAME\_EVENT}}, in which anchoral edges are removed, assessing the impact of event grouping on version selection. Table~\ref{tab:ablation} presents the results.}

\begin{table}[ht!]
\centering
\small
\setlength{\tabcolsep}{3pt}
\begin{tabular}{|l|c|c|c|}
\hline
\textbf{Metric} & \textbf{w/o \texttt{BEFORE}} & \textbf{w/o \texttt{SAME\_EV.}} & \textbf{TAEG} \\
\hline
ROUGE-1 F1   & \rev{0.886} & \rev{0.929} & \rev{0.918} \\
ROUGE-2 F1   & \rev{0.811} & \rev{0.865} & \rev{0.848} \\
ROUGE-L F1   & \rev{0.809} & \rev{0.866} & \rev{0.846} \\
BERTScore F1 & \rev{0.935} & \rev{0.964} & \rev{0.906} \\
METEOR       & \rev{0.469} & \rev{0.575} & \rev{0.550} \\
\hline
\rev{Selection acc.} & \rev{0.341} & \rev{0.489} & \rev{0.489} \\
\hline
\end{tabular}
\caption{\rev{Ablation of the TAEG's edge types. Both variants keep the final timeline iteration, so all configurations preserve $\tau = 1.000$ by design. Selection accuracy is the oracle selection accuracy of Section~\ref{sec:selection_results} (random floor $\approx 0.37$).}}
\label{tab:ablation}
\end{table}

\revii{Before interpreting the results, it is important to delimit what each ablation removes. Both variants remove \emph{edges from the graph over which centrality is computed}; neither removes the event grouping itself, which is not a component of the TAEG but part of the task input --- the alignment between event versions and canonical events comes from the chronology, and it is what defines the candidate set of each event in line~6 of Algorithm~\ref{alg:taeg}. Removing the grouping would not ablate a design choice: with no candidate set, the selection step would maximize a fixed global score over the whole graph and emit the same version at every position, producing a degenerate output. The ablations therefore answer a well-posed question --- what does each relation contribute \emph{as weighted evidence for centrality} --- and not whether event clustering is needed.}

\revii{With that scope in mind, the results give a clear and partly unexpected picture. Removing the \texttt{BEFORE} edges collapses selection quality to the random floor (oracle accuracy 0.341; ROUGE-L 0.809, indistinguishable from Timeline+Random): the discriminative signal exploited by centrality flows through the narrative-flow similarities encoded by the temporal edges. Removing the \texttt{SAME\_EVENT} edges, in contrast, leaves oracle accuracy unchanged at 0.489 and slightly improves the corpus metrics. The coincidence of the accuracy figures does not mean that the two configurations behave identically: their outputs differ, as the divergent corpus metrics and output lengths in Table~\ref{tab:ablation} show, so a number of per-event choices do change --- they simply trade correct and incorrect selections in equal measure.}

\revii{The explanation is that parallel accounts of the same event are nearly equidistant in the TF-IDF space: any version resembles its siblings about as much as any other, so intra-cluster weights are close to uniform and carry almost no information for ranking. This is a negative result, and we report it as such, because it is informative for the design of future systems: it indicates that a graph over event versions should not expect intra-event lexical similarity to discriminate between candidates, and that models building on this structure --- including graph neural approaches --- will need weights derived from semantic, discourse, or factuality features rather than from surface overlap. The corollary for the TAEG is equally direct: of its two relations, only the temporal one earns its place as evidence for selection.}

\subsection{\rev{Sensitivity to Incomplete Timelines}}
\label{sec:degradation}
\rev{Real-world chronologies are rarely complete. To provide a first quantification of the framework's robustness to incomplete temporal knowledge, we randomly remove a percentage of the canonical events from the timeline before building the TAEG, and measure the impact on all metrics. Table~\ref{tab:degradation} reports the results, averaged over 10 runs per degradation level.}

\begin{table}[ht!]
\centering
\small
\setlength{\tabcolsep}{2pt}
\begin{tabular}{|l|c|c|c|c|}
\hline
\textbf{Metric} & \textbf{0\%} & \textbf{10\%} & \textbf{25\%} & \textbf{50\%} \\
\hline
ROUGE-L F1   & \rev{0.846} & \rev{0.795 {\tiny$\pm$.022}} & \rev{0.723 {\tiny$\pm$.037}} & \rev{0.534 {\tiny$\pm$.040}} \\
BERTScore F1 & \rev{0.906} & \rev{0.897 {\tiny$\pm$.015}} & \rev{0.882 {\tiny$\pm$.021}} & \rev{0.878 {\tiny$\pm$.033}} \\
METEOR       & \rev{0.550} & \rev{0.474 {\tiny$\pm$.021}} & \rev{0.397 {\tiny$\pm$.027}} & \rev{0.246 {\tiny$\pm$.026}} \\
\hline
\end{tabular}
\caption{\rev{Impact of randomly removing canonical events from the timeline (degradation level in \% of events removed; mean $\pm$ std over 10 runs). Removed events are absent from the output; among the remaining events, ordering stays correct by design.}}
\label{tab:degradation}
\end{table}

\rev{Content metrics degrade gracefully and roughly in proportion to the lost coverage, confirming that the impact of an incomplete timeline is completeness-driven rather than disruptive: among the remaining events, ordering and selection are unaffected. Notably, even with half of the timeline removed, the consolidation retains a ROUGE-L F1 of 0.534 --- still far above the timeline-agnostic LexRank baseline (0.206) operating with full access to the texts. A partial temporal backbone is thus considerably better than none, a result that establishes the robustness reference that automatic (and hence imperfect) event-alignment methods will have to meet.}

\subsection{\rev{Selection-Level Evaluation}}
\label{sec:selection_results}
\rev{Table~\ref{tab:selection} reports the oracle selection accuracy of Section~\ref{sec:selection_metric}, computed over the 88 contested events for which a golden segment could be reliably aligned (out of 96 contested events; the analytical random floor on this set is $\approx 0.35$, empirically $0.380$). This selection-level view discriminates the strategies far more sharply than the (saturating) corpus metrics.}

\begin{table}[ht!]
\centering
\small
\setlength{\tabcolsep}{4pt}
\begin{tabular}{|l|c|}
\hline
\textbf{Strategy} & \textbf{Oracle selection accuracy} \\
\hline
Timeline+Random    & \rev{0.372 {\tiny$\pm$ 0.052}} \\
Timeline+Priority  & \rev{0.409} \\
Timeline+Centroid  & \rev{0.375} \\
Timeline+Longest   & \rev{\textbf{0.648}} \\
TAEG (Algorithm~\ref{alg:taeg}) & \rev{0.489} \\
TAEG w/o \texttt{BEFORE}      & \rev{0.341} \\
TAEG w/o \texttt{SAME\_EVENT} & \rev{0.489} \\
\hline
\end{tabular}
\caption{\rev{Oracle selection accuracy on the 88 evaluated contested events (oracle = version with highest ROUGE-L F1 against the event's golden segment; random floor $\approx 0.37$).}}
\label{tab:selection}
\end{table}

\rev{Three observations follow. First, the TAEG's centrality-based selection (0.489) is clearly and significantly above the random floor: across 30 seeded random runs, it sits at the 100th percentile of the random distribution for oracle accuracy, ROUGE-1/2/L, and METEOR --- the graph carries a genuine selection signal. Second, the purely local Timeline+Centroid baseline performs at the floor (0.375), showing that intra-cluster similarity alone is uninformative (parallel accounts are nearly equidistant) and that the TAEG's advantage over it comes from the global graph structure. Third, Timeline+Longest reaches 0.648: with a fusion-style reference over complementary sources, the most detailed version is most often the best available one, making length a remarkably strong proxy \emph{on this benchmark}. We report this candidly: it characterizes the benchmark as much as the methods, and it sharpens the research question for noisier domains, where detail and faithfulness no longer coincide.}

\section{\revii{Findings and Implications for the Task}}
\label{sec:findings}
\revii{Benchmarking the reference systems produces four findings. Because this study aims to establish the task rather than to advocate a method, we state them as properties of Narrative Consolidation and of the resource, and draw out what each implies for future work.}

\revii{\textbf{Finding 1: chronological structure, not content selection, is the defining difficulty.} The gap between timeline-agnostic and timeline-aware systems (ROUGE-L F1 0.206 vs.\ 0.811--0.947) dwarfs the spread among selection criteria. This is the empirical justification for treating Narrative Consolidation as a task distinct from Multi-Document Summarization: a strong extractive summarizer applied without chronological structure does not merely score lower, it produces an output of a different kind, in which the sequence of events is lost (Section~\ref{sec:baseline_lexrank}). It also implies that progress on this task will come primarily from how chronology is obtained and represented --- and, since the timeline is an input here, that the induction of chronological structure is the critical open problem (Section~\ref{sec:limitations}).}

\revii{\textbf{Finding 2: a length heuristic is a strong reference point on fusion-style references.} Selecting the longest available version per event attains 0.947 ROUGE-L F1 and 0.648 oracle selection accuracy, ahead of every other system tested. Simple heuristics that resist more elaborate alternatives are a familiar phenomenon in summarization, where lead-based baselines long remained competitive with sophisticated models; the analogous observation for this task is that when the reference consolidation fuses complementary, non-conflicting accounts, the detail of an account and its quality nearly coincide. Two implications follow. For evaluation: any future system on this resource must be compared against a length heuristic, not only against a random or centrality-based one, and we supply it for that purpose. For resource design: benchmarks in which the sources genuinely conflict are needed to separate detail from faithfulness, which is why we identify a news-domain resource as a priority (Section~\ref{sec:limitations}).}

\revii{\textbf{Finding 3: an explicit relational prior yields signal, but not enough.} Centrality over the TAEG selects the oracle version in 48.9\% of contested events, against a random floor of about 37\% and at the 100th percentile of the seeded random distribution --- the structure carries genuine information --- yet it remains below the length heuristic. Structural priors, therefore, are not automatically superior to trivial ones on this task; the burden on a structured method is not to beat chance but to beat completeness. We state this as the concrete open challenge posed by the benchmark.}

\revii{\textbf{Finding 4: intra-event lexical similarity is uninformative for version selection.} The ablation of Section~\ref{sec:ablation} shows that the discriminative signal lies in the temporal relations, while similarity weights among versions of the same event are effectively uniform. For future models --- in particular graph neural architectures over this structure --- the implication is that useful intra-event edge weights must encode semantic, discourse, or factuality distinctions rather than surface overlap. Reporting this negative result is, in our view, one of the more useful outcomes of the benchmark, since it removes a design avenue that would otherwise appear natural.}

\section{Limitations and Future Work}
\label{sec:limitations}
While our results are promising, we acknowledge several limitations\rev{, which we present as an explicit research roadmap}.

\rev{First, \revii{this study assumes} an external, pre-defined timeline. As discussed in Section~\ref{sec:formal_definition}, this was a deliberate scoping decision \revii{that isolates the version-selection subproblem} under a known chronology. Automatically inducing the timeline from the documents themselves requires solving cross-document event extraction, cross-document event coreference, and temporal ordering --- each an open research problem in its own right --- and is precisely the focus of the authors' ongoing work, as the next stage of this research program. The sensitivity analysis of Section~\ref{sec:degradation} provides the robustness reference that such automatic (and hence imperfect) alignment methods will have to meet. Handling \emph{contradictory} (rather than merely incomplete) timelines remains open.}

Second, our evaluation is based on a single, albeit complex, dataset from the religious domain\rev{, which, as discussed in Section~\ref{sec:contradictions}, deliberately controls for factual contradictions and adversarial perspectives}. \revii{Testing the task and its reference systems} on other multi-perspective narrative corpora, such as legal depositions or journalistic reports, is crucial to assess its generalizability. \rev{To the best of our knowledge, however, no existing corpus simultaneously provides multiple overlapping accounts, an aligned canonical timeline, and a reference consolidation; constructing such a benchmark for the news domain --- including annotations of contradictions and perspectives --- is a substantial research contribution that we leave as future work.}

\rev{Third, regarding stronger baselines from the Timeline Summarization literature: off-the-shelf TLS/MDTS systems are date-driven, relying on explicit temporal expressions and document timestamps to anchor content, and producing lists of dated summaries rather than a single continuous narrative. Since our corpus contains no explicit dates and the task's target output is a fluid narrative, a direct comparison would be ill-defined. The timeline-aware baselines of Section~\ref{sec:fair_baselines} play this role instead, granting competing methods the same temporal knowledge in a format compatible with the task; adapting TLS methods to Narrative Consolidation remains an interesting direction.}

\rev{Finally, the natural extension of this work is abstractive: leveraging the TAEG as a structural backbone --- e.g., with a GNN encoder \citep{yasunaga2017graph} and an abstractive decoder \citep{xiao2022primera, zhang2020pegasus} --- so that the versions within each \texttt{SAME\_EVENT} cluster are fused into enriched event descriptions. A first investigation of abstractive Narrative Consolidation on this same resource is presented in a companion study by the authors \citep{Finger:2026}.}

\subsection{\revii{Influence of Genre, Domain, and Language}}
\label{sec:generalizability}
\revii{Since the present findings rest on a single resource, we note how they depend on its properties. \textbf{Genre} governs the relationship between detail and faithfulness, and therefore the strength of the length heuristic (Finding 2): the Gospels are complementary testimonial accounts with stable content and no adversarial intent, so a longer account is usually a more informative one, whereas in breaking news, where successive reports revise earlier facts, and in legal depositions, where accounts are strategically framed, we would expect detail to decouple from faithfulness, increasing the value of selection mechanisms that model contradiction and stance (Section~\ref{sec:contradictions}). Genre also determines what temporal evidence is available: date-wise timeline summarization presupposes explicit temporal expressions and document timestamps \citep{Martschat:2017, Mamidala:2021}, and dedicated temporal reasoning is required where such signals are implicit or noisy \citep{Song:2025}, whereas the narrative testimony studied here supplies neither. \textbf{Domain} determines whether a canonical timeline exists at all: our assumption is realistic for domains with an established chronology --- scriptural harmonies, case files, clinical histories --- but not for open-domain collections, where the chronology must be induced, and the degradation analysis of Section~\ref{sec:degradation} covers missing events but not \emph{erroneous} ordering, the more likely failure mode of an induced timeline. \textbf{Language} affects the resource and the metrics differently. The resource is largely language-independent, since the alignment is expressed over \texttt{book:chapter:verse} references, so the same 169-event chronology applies to any translation and parallel benchmarks can be built at low cost. The metrics are not: ROUGE matches surface $n$-grams \citep{lin2004rouge}, so morphological variation understates agreement; METEOR depends on stemming and synonymy modules that must be available for the target language \citep{banerjee2005meteor}; and BERTScore inherits the quality of the contextual embeddings trained for it \citep{zhang2020bertscore}. Establishing how these proxies behave across languages is thus a prerequisite for cross-lingual replication, which we have not undertaken.}

\section{Conclusion}
This paper introduced and formalized \emph{Narrative Consolidation}, a new NLP task that reframes the processing of multiple overlapping narratives, arguing for the primacy of coherence and completeness over simple compression. \revii{Around that definition we assembled what the task needs in order to be studied: an evaluation paradigm that includes a selection-level metric, the Gospel Consolidation Language Resource, and a graded suite of reference systems spanning timeline-agnostic centrality, timeline-aware heuristics, and the Temporal Alignment Event Graph, together with ablations and a robustness analysis.}

\revii{Benchmarking these systems characterizes the task more sharply than any of them individually. Chronological structure, rather than content-selection sophistication, is the dominant factor: granting a canonical timeline moves ROUGE-L F1 from 0.206 to at least 0.811, while the criterion used to choose among versions accounts for the remaining margin. Within that margin, an explicit relational structure yields a real but modest signal, and is outperformed on this resource by a simple length heuristic --- an outcome that says as much about fusion-style references as about the systems, and that we therefore record as a property of the benchmark. The ablations add a negative result of direct use to future designs: the discriminative evidence lies in the temporal relations, whereas lexical similarity among versions of the same event is uninformative.}

\revii{We consequently offer the TAEG not as a solution to Narrative Consolidation but as one reference point among several, and we leave a concrete challenge in its place: on this benchmark, no principled selection mechanism has yet surpassed selecting the longest account; doing so --- particularly on resources where sources genuinely conflict --- and moving beyond selection to the fusion of complementary details are the natural next objectives. Alongside it, the induction of the chronological backbone without an external timeline remains, in our assessment, the decisive open problem for the task. The resource, the reference implementations, and the evaluation code are publicly available so that both challenges can be taken up directly.}

\section*{Declarations}

RF is the main contributor and writer of this manuscript. EC and GR reviewed the manuscript. All authors read and approved the final manuscript.

The authors declare that they have no competing interests.

This work was partially supported by CNPq (grant 313845/2023-9).

The datasets and software generated and analysed during the current study are available at https://github.com/neemias8/TAEG.

\section{Bibliographical References}\label{sec:reference}

\bibliographystyle{apalike-sol}
\bibliography{refs}

\end{document}